\documentclass[11pt,a4paper]{article}
\usepackage[hyperref]{acl2018}
\usepackage{times}
\usepackage{latexsym}
\usepackage{amsmath}
\usepackage{amsfonts}
\usepackage{multirow,array}

\usepackage{url}

\usepackage{graphicx}
\usepackage{subcaption}

\aclfinalcopy 
\def\aclpaperid{13} 

\newcommand*\samethanks[1][\value{footnote}]{\footnotemark[#1]}

\hypersetup{draft} 
\begin{document}

\title{Quantum-inspired Complex Word Embedding}

\author{Qiuchi Li\Thanks{ Corresponding author}\\
  University of Padova \\
  Padova, Italy \\
  {\tt qiuchili@dei.unipd.it}\\
  \And
  Sagar Uprety\samethanks\\
  The Open University \\
  Milton Keynes, UK\\
  {\tt sagar.uprety@open.ac.uk } \\ 
  \AND
  Benyou Wang \\
  University of Padova \\
  Padova, Italy\\
  {\tt wabyking@163.com } \And
  Dawei Song \\
  The Open University \\
  Milton Keynes, UK\\
  Beijing Institute of Technology\\
  Beijing, China\\
  {\tt dawei.song@open.ac.uk}
  }

\date{}

\maketitle

\begin{abstract}
A challenging task for word embeddings is to capture the emergent meaning or polarity of a combination of individual words. For example, existing approaches in word embeddings will assign high probabilities to the words "Penguin" and "Fly" if they frequently co-occur, but it fails to capture the fact that they occur in an opposite sense - Penguins do not fly. We hypothesize that humans do not associate a single polarity or sentiment to each word. The word contributes to the overall polarity of a combination of words depending upon which other words it is combined with. This is analogous to the behavior of microscopic particles which exist in all possible states at the same time and interfere with each other to give rise to new states depending upon their relative phases. We make use of the Hilbert Space representation of such particles in Quantum Mechanics where we subscribe a relative phase to each word, which is a complex number, and investigate two such quantum inspired models to derive the meaning of a combination of words. The proposed models~\footnote{https://github.com/complexembedding/ \\ complex$\_$word$\_$embedding.git} achieve better performances than state-of-the-art non-quantum models on the binary sentence classification task.
\end{abstract}

\section{Introduction}

Word embeddings~\cite{Bengio:2003:NPL:944919.944966,Mikolov:2013:DRW:2999792.2999959,pennington2014glove} are the current state of art techniques to form semantic representations of words based on their contexts. They have been successfully used in various downstream tasks such as text classification, text generation, etc. Building on word embeddings, various unsupervised~\cite{Kiros:2015:SV:2969442.2969607,DBLP:journals/corr/HillCK16} and supervised~\cite{D17-1070} models for sentence embeddings have been proposed. The general idea behind word embeddings is to use word co-occurrence as the basis of semantic relationship between words. This naturally brings about the difficulty for word embedding approaches in capturing the emergent meaning of a combination of words, such as a phrase or a sentence. For example, the phrase "ivory tower" can hardly be modeled as a semantic combination of "ivory" and "tower". Or, the high frequency of occurrence of the words "Penguin" and "Fly" fails to suggest that they are negative correlated. 

In the field of information retrieval (IR), various models based on the mathematical framework of Quantum Theory have been applied to capture and represent dependencies between words~\cite{sordoni_modeling_2013,xie_modeling_2015,zhang_end_to_end_2018}, inspired by the pioneering work of~\citet{van2004geometry}. ~\citet{sordoni_modeling_2013} models a segment of text as a quantum mixed state, represented by a positive semi-definite matrix called density matrix in a Hilbert Space, whose non-diagonal entries entail word relations in a quantum manner(Quantum Interference). The resulting Quantum Language Model (QLM) outperforms various classical models on ad-hoc retrieval tasks. ~\citet{xie_modeling_2015} captures Unconditional Pure Dependence (UPD)~\cite{hou_mining_2013} between words in a quantum way by demonstrating the equivalence relation between UPD and Quantum Entanglement (QE) and providing a way to incorporate UPD information into QLM, leading to improved performance over the original QLM. ~\citet{zhang_end_to_end_2018} develops a well-performing question answering (QA) system by extracting various features and learning to compare the density matrices between a question and an answer.

The successful application of quantum-inspired models onto IR tasks~\cite{wang2016exploration} to some extent demonstrates the non-classical nature of word dependency relations. However, all these models simplify the space of interest to be space of real vectors $~\mathbb{R}^n$, with the representation of a word or a text segment being a real-valued vector or matrix, largely due to the lack of proper textual features corresponding to the imaginary part. Since quantum phenomena cannot be faithfully expressed without complex numbers, these models are theoretically limited. In a recent work,~\citet{aerts_towards_2017} presents a theoretical quantum framework for modeling a collection of documents called QWeb, in which a concept is represented as a state in a Hilbert Space, and concept combination is represented as a superposition of the concept states. Under this framework, the complex phases of each concept have a natural correspondence to the extent of interference between concepts. However, the framework has not given rise to any applicable models onto IR or NLP tasks to the authors' knowledge. 

Inspired by the potential of quantum-inspired models to represent word relations, we seek to build quantum models to represent words and word combinations, and explore the use of complex numbers in the modeling process. Our model is built on top of two hypothesis: I) A word is a linear combination of latent concepts with complex weights. II) A combination of words is viewed as a complex combination of word states, either a superposition state or a mixed state.  The first hypothesis agrees with QWeb, but here we concretize a concept in QWeb to be a word. The second hypothesis is an extension of both QWeb and the work by ~\citet{zhang_end_to_end_2018}, because QWeb restricts a combination of concepts to be a superposition state while the work by ~\citet{zhang_end_to_end_2018} assumes that a sentence is a complex mixture of word projectors.

This study sets foot in sentence-level analysis, and treats a sentence as a combination of words. We intend to model a word as a quantum state containing two parts: amplitudes and complex phases, and expect to capture the low-level word co-occurrence information by the amplitudes, while using the phases to represent the emergent meaning or polarity when a word is combined with other words. We investigate on two models to represent the combination of words, either as a superposition of word states or as a mixture of word projectors. The effectiveness of the two models are evaluated on 5 benchmarking binary sentence classification datasets, and the results show that the mixture model outperforms state-of-the-art word embedding approaches.

The motivation behind this paper stems from an analogy with Quantum Physics. Consider the phrase "Penguins fly". If we model it along the lines of the famous double slit experiment in Quantum Physics, the two slits corresponds to human interpretation of words "Penguins" and "Fly" (Verb sense of Fly). When only one slit is open at a time, the waves corresponding to the individual word will go through the slit and register onto the screen.  The screen is made of a set of polarity detectors judging opinion or sentiment polarities.

In Figure 1.a the human mind sees the word 'Penguins' alone and detects it as a neutral word with a very high probability. This is analogous to the double slit experiment with one slit open. The same is the case for the word 'Fly' considered in isolation. By classical logic, when the two words are taken together as a phrase 'Penguins fly', the human mind should assign a high probability of it being neutral again. However, we know that it is a false statement (Figure 1.c). 

\begin{figure*}[h!]
   \begin{subfigure}[H]{0.5\textwidth}
        \includegraphics[width=\textwidth]{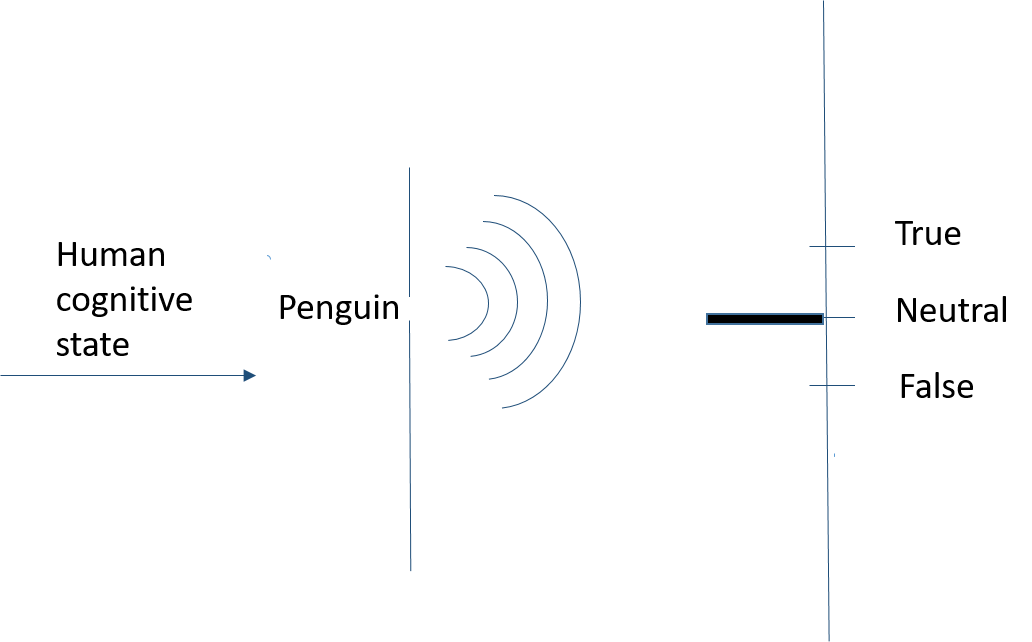}
        \caption{Figure 1.a}
   \end{subfigure}
    \begin{subfigure}[H]{0.5\textwidth}
        \includegraphics[width=\textwidth]{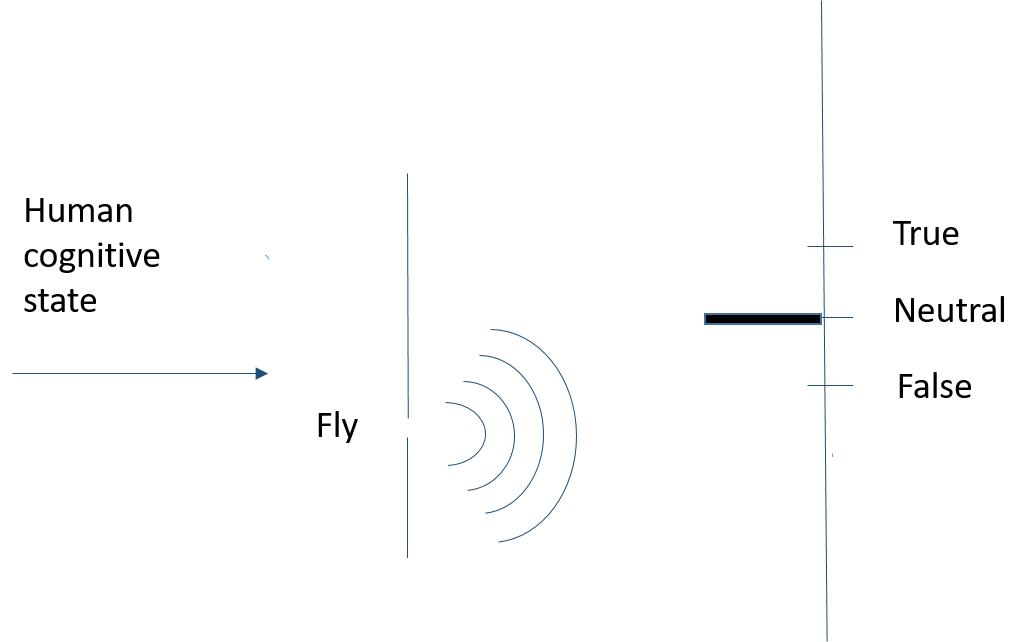}
        \caption{Figure 1.b}
    \end{subfigure}

    \begin{subfigure}[H]{0.5\textwidth}
        \includegraphics[width=\textwidth]{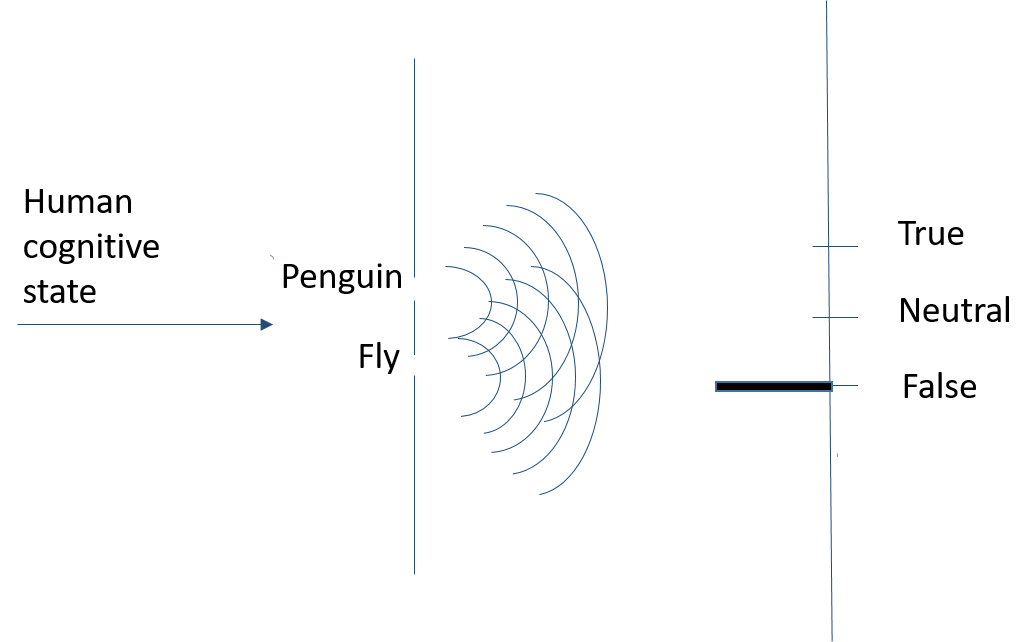}
        \caption{Figure 1.c}
    \end{subfigure}
     
\end{figure*}

Different from classical representation, this study hypothesizes that the combination of words can be viewed as a superposition or complex mixture of quantum entities which gives rise to a new state. In this way, the emerging meaning or polarity of a combination of words will manifest in the interference between words, and be captured inherently in the density matrix representation. For example, two or more words having a neutral sense individually may combine to give a negative sense, just like the case in the analogy given above.

\section{Hilbert Space Representation of Words and Sentences}

The section introduces the proposed quantum framework for representing words and sentences. Our research scope is currently limited to sentence and word level analysis. However, our proposed model is potentially capable of representing higher-level concepts such as paragraphs and documents, which we will investigate in the future. To be consistent with the quantum framework, we use Dirac notations, in which a unit vector $\vec{\mu}$ and its transpose $\vec{\mu}^T$ are denoted as ket $\vert u \rangle$ and bra $\langle u \vert$ respectively. 

Suppose there are $n$ independent latent concepts in the text collection, we then model words and sentences as quantum concepts defined on an $n$-dimensional Hilbert Space $\mathbb{H}^n$, where latent concepts form a set of pure orthonormal states of the space. Using Dirac notations, the concepts are denoted as $\{\vert C_i \rangle\}_{i=1}^n$. Intuitively, latent concepts correspond to the contexts in which words are used.

Each word $t$ is modeled as a superposition state~\cite{Nielsen:2011:QCQ:1972505} in the $n$-dimensional Hilbert Space $\mathbb{H}^n$. Equivalently, it can be viewed as a linear combination of $\{\vert C_i \rangle\}_{i=1}^n$ with complex weights, i.e. $\vert t \rangle =  \sum_{k=1}^n e^{i\theta_k}w_k \vert C_k \rangle$, in which $\{w_i\}_{i=1}^n$ are real-valued amplitudes with $w_i>0$ and $\sum_{i=1}^n w_i^2 =1$, and $\theta_i \in [-\pi,\pi], i=1,2,...,n$ are the corresponding complex phases. This representation can be seen as a generalization of previous word embedding approaches~\cite{Bengio:2003:NPL:944919.944966,Mikolov:2013:DRW:2999792.2999959,pennington2014glove} in that it can be regarded as a complex embedding with unitary length of word vectors. A word has many different contexts associated with it. For example, 'Penguin' is associated with 'Bird', 'Antarctica', 'Snow', etc. When a quantum particle(e.g. electron) is said to be in a superposition state, it exists in a new state(e.g. position) of all of its possible outcomes(at all positions) at the same time. A particular outcome is observed upon measurement. Similarly a word exists in all of its contexts at the same time and depending upon its interaction with other words in a combination, a particular context is materialized. Note that because of reduced dimensionality, the contexts are latent concepts.

A sentence is a non-classical combination of words. Since each word is a superposition of latent concepts, a sentence $s$ is also a non-classical combination of latent concepts $\{\vert C_i \rangle\}_{i=1}^n$. It is represented by a $n$ by $n$ density matrix $\rho$ which is positive semi-definite with unitary trace: $\rho \geq 0$, Tr($\rho$) = 1. The real diagonal values of $\rho$ reflects the strength of concepts in the sentence, whereas the non-diagonal values encodes correlations between concepts in a quantum manner. The density matrix can be computed from the word states either directly or through a training strategy.   

Our proposed approach is related to but largely differs from~\citet{sordoni_modeling_2013} and~\citet{zhang_end_to_end_2018}.~\citet{sordoni_modeling_2013} models queries and documents as density matrices and provides a training method for constructing density matrices from texts.~\citet{zhang_end_to_end_2018} directly computes the density matrix of a sentence and put it into an end-to-end neural network for handling the Question Answering (QA) task. Both works view a segment of texts as a mixed state~\cite{Nielsen:2011:QCQ:1972505} and use real-valued density matrix as a representation. Our study also directly computes the sentence representation from the word superposition states. However, different from both works, our study explores on treating a sentence as either a strictly mixed state or a superposition state. In either case, it can be represented as a complex density matrix with complex values for non-diagonal entries.

On top of the obtained sentence representation, different quantum operations can be applied to achieve a particular NLP target at hand. For sentence classification tasks, one can perform projective measurements onto the sentence representation to determine the sentiment polarity; for sentence text similarity task, the amplitude of the inner product between a sentence pair may provide evidence for judging to what extent they are similar to each other. Projective measurements and inner products are methods to compute probabilities in Quantum Theory~\cite{Nielsen:2011:QCQ:1972505}. 

\section{Complex Embedding Network for Text Classification}

In this paper, we build a complex embedding network for text classification on the basis of Hilbert Space representation for words and sentences. The end-to-end network accepts a sentence sequence as input and computes its classification label in the procedure shown by Figure 2:

\begin{figure*}[t]\label{model}
  \caption{The process diagram of the proposed complex embedding network. $\vert V \vert$ is the vocabulary size, n is the embedding dimension, m is the maximum length of a sentence}
\includegraphics[width=\textwidth]{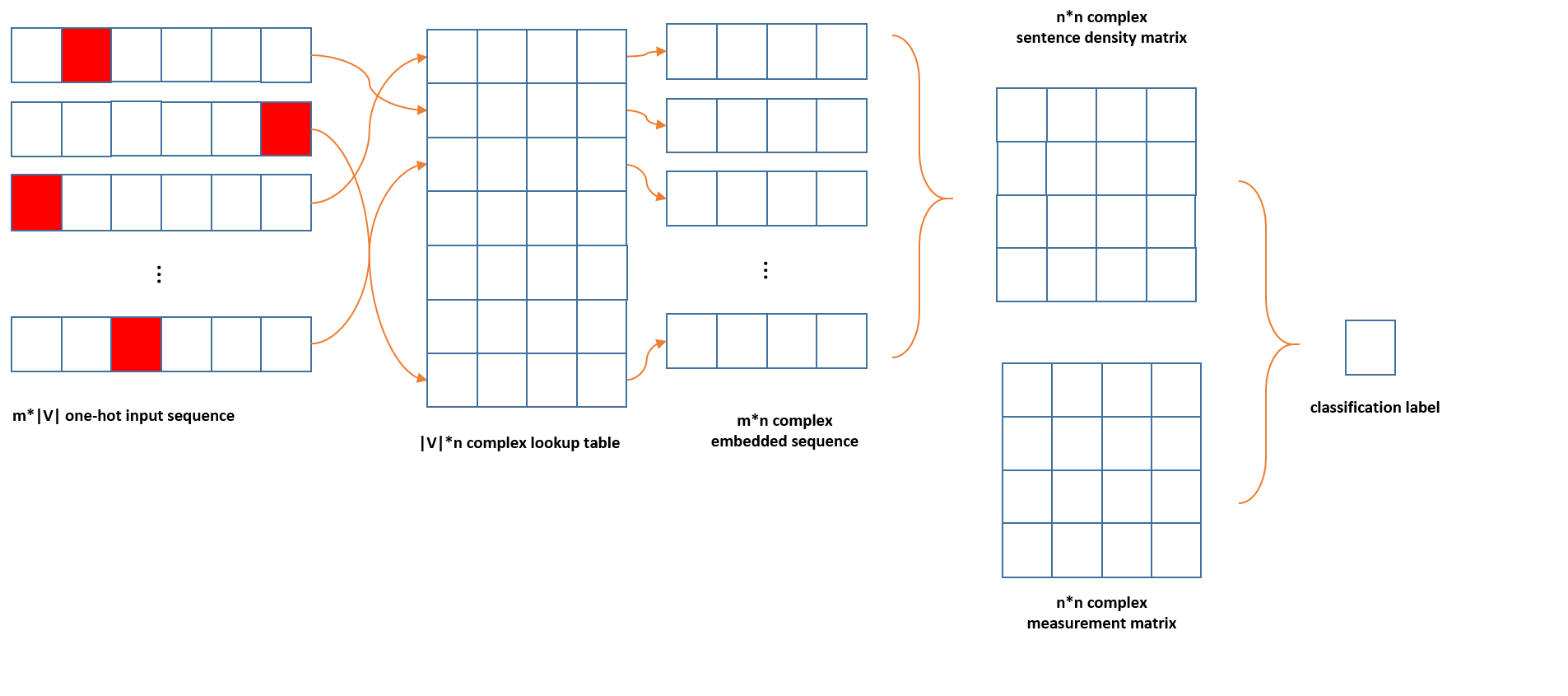}
\end{figure*} 

The input one-hot sequence is passed through an embedding layer with a complex valued lookup table, which maps each word into a complex vector representing its superposition state, resulting in a sequence of complex embedding vectors. Then the density matrix of the sentence is computed from the complex embedding vectors. Finally, a square projection matrix takes control of the measurement. For any sentence state $\rho$, the measurement probability is computed through Born's rule~\cite{born_zur_1926}:
	
\begin{equation} \label{born}
\begin{split}
	p =  Tr(P \rho)
\end{split}
\end{equation}

Where $P$ is a projection matrix satisfying $P^2 = P, P = P^T$. The value of $p$ determines the class of this sentence. The lookup table determining the complex embedding for each word is learned by feeding the network with a sufficient number of training data.

The crucial step of the process falls on how to compute the sentence density matrix from the sequence of complex word embeddings. As no previous research has attempted to build complex networks for text classification task, we investigate on two approaches for this step:

I) A sentence is viewed as a linear combination of all word vectors in the sentence, i.e. $ \vert S \rangle =  \frac{\sum_{l=1}^m {\lambda_l \vert t_l \rangle}}{||\sum_{l=1}^m {\lambda_l \vert t_l \rangle}||_2} $, with $\sum_{l=1}^m \lambda_l =1$. Here $\lambda_l$s are real-valued weights indicating the relative degree of importance for each word in the sentence, and the state is divided by its 2-norm in order to guarantee it is a legal quantum state (i.e.,with unit length). The sentence is then a pure superposition state and the density matrix can be computed simply as $\rho = \vert S \rangle \langle S \vert$.

II) A sentence is viewed as a classical mixture of the word states in the sentence, i.e.  $ \rho =  \sum_{l=1}^m {\lambda_l \vert t_l \rangle \langle t_l \vert} $, with $\sum_{l=1}^m \lambda_l =1$. Here $\vert t_l \rangle \langle t_l \vert$ is the density matrix representing the superposition state of a word $t_l$. This equation guarantees the obtained $\rho$ is a legal density matrix without any further normalization.

The constructed density matrix representing a sentence has real values for diagonal entries and non-zero complex values for non-diagonal entries. Intuitively, the diagonal entries tell us something about the distribution of latent concepts in the sentence, whereas the non-diagonal values entail information regarding the emergent meanings. Consider a very simple example where the complex phases represent positive, neutral or negative senses. Independently, both the words "Penguin" and "Fly" have neutral sense, $\theta_P = \theta_F = 0$. When they are combined together in a sentence, then sentence density matrix has a negative-phased complex value in the entry corresponding to them, i.e. $\theta_{PF} < 0$. Therefore, the combination of these two words will have a negative complex phase, implying the negative sense "Penguins cannot fly". In practice, the connections between words are much more complicated, but we believe that by feeding the above-mentioned models with enough data, the constructed density matrix will be able to effectively capture and represent the emergent meanings of sentences.

The above-mentioned approaches lead to two different models, resulting in different embeddings learned from the same training data. Hence we name them as complex embedding superposition  (CE-Sup) network and complex embedding mixture (CE-Mix) network respectively. For sake of simplicity, we assign equal importance of each word in the sentence representation in both models, i.e. $\lambda_l = \frac{1}{m}, l = 1,2,...,m$. In a relevant research,  ~\citet{zhang_end_to_end_2018} learns the values of $\lambda_l$s in the training framework, while enforcing the word embeddings $\vert t_l \rangle$s to be fixed. By fixing $\lambda_l$s and learning $\vert t_l \rangle$s from the data, this paper is essentially aiming at obtaining better representation of each word from the training data, whereas Zhang et al.'s work directly takes existing word vectors trained from external corpus. It would be interesting to see what a co-training of $\vert t_l \rangle$s and $\lambda_l$s will bring about in future works.

\section{Experimental Setup}

The experiments are conducted on five benchmarking datasets for binary text classification: Customer Review dataset (CR)~\cite{hu_mining_2004}, Opinion polarity dataset (MPQA)~\cite{wiebe_annotating_2005}, Sentence Subjectivity dataset (SUBJ)~\cite{pang_seeing_2005}, Movie Review dataset (MR)~\cite{pang_seeing_2005}, and Stanford Sentiment Treebank (SST) dataset~\footnote{https://nlp.stanford.edu/sentiment/index.html}. The statistics for the datasets are shown in Table 1.

\begin{table}[ht]
\begin{center}
\caption{Dataset Statistics}\label{table:2}
\begin{tabular}{l|llll}
 \hline
  Dataset & \#Count & Task & Classes\\ \hline
 CR & 4k & product reviews & pos/neg \\
 MPQA & 11k & opinion polarity & pos/neg\\ 
 SUBJ & 10k & subjectivity & subj/obj \\
 MR & 11k & movie reviews & pos/neg  \\ 
 SST & 70k & movie reviews & pos/neg \\ 
 \hline
\end{tabular}
\end{center}
\end{table}

In this paper, we compare the classification accuracy of our proposed Complex Embedding Superposition (CE-Sup) network and Complex Embedding Mixture (CE-Mix) network with three existing unsupervised representation training models, Unigram-TFIDF and fastText Bag-of-Words (BOW), as well as two existing supervised representation training models, namely CaptionRep BOW~\cite{hill_learning_2016_a} and DictRep BOW~\cite{hill_learning_2016_b}.  We directly take the performances of these systems on the 5 datasets from existing works. Since the performances for CaptionRep and DictRep are not available on SST, we use the performance of another model called Paragraph-Phrase~\cite{bansal_towards_2016}. For a fair comparison, we also implement an end-to-end supervised real embedding network (Real-Embed), where each word is mapped to a real-valued vector in the embedding layer, based on which the sentence representation is obtained by averaging the embedding vectors for all words in the sentence, and a fully connected layer maps the sentence vector to the classification label. CE-mixture, CE-Superposition and Real-Embed are trained and tested in a completely identical process. 

For the construction of training, validation and test data, they are readily available for SST dataset, and for the other four datasets we randomly split the whole data into 8:1:1 for training, validation and test data respectively. The embedding dimension is set to be 100. We use batch training with batch size being 32 for SST and 16 for the other datasets. We adopt Adam as the optimizer and use the default parameters for Adam in Keras~\footnote{https://keras.io/}.

The experiments are implemented in Keras and Tensorflow~\footnote{https://www.tensorflow.org/} under Python 3.6.4. The experiment is run on a desktop with NVidia Quadro M4000 and 16GB RAM. 

\section{Results and Discussion}
\begin{table*}[ht]
\begin{center}
\caption{Experimental Results in percentage(\%). The best performed value for each dataset is in bold.}\label{table:3}
\begin{tabular}{p{4cm}<{\centering}|rrrrr}
 \hline
  \textbf{Model}& \textbf{CR} & \textbf{MPQA }& \textbf{MR} & \textbf{SST} & \textbf{SUBJ} \\ \hline
  Unigram-TFIDF & 79.2& 82.4 & 73.7 & - &90.3\\ 
  word2vec BOW & 79.8 & \textbf{88.3} & 77.7 & 79.7 &90.9 \\
  fastText BOW & 78.9& 87.4 & 76.5 & 78.8 &91.6\\ \hline
  CaptionRep BOW & 69.3& 70.8 & 61.9 & - &77.4\\ 
  DictRep BOW & 78.7& 87.2 & 76.7 & - &90.7\\ 
  Paragram-Phrase & - & - & - & 79.7 & -\\ \hline
  Real-Embed & 77.5& 84.7 & 77.0 & 80.0 &92.0\\ 
  CE-Sup & 80.0& 85.7 & 78.4 & 82.6 &92.6\\ 
  CE-Mix & \textbf{81.1}& 86.6 & \textbf{79.8} & \textbf{83.3} &\textbf{92.8}\\ \hline
 
\end{tabular}
\end{center}
\end{table*}

In this study, we seek to answer the following two research questions:
\begin{itemize}
\item[\textbf{RQ1.}] Do the proposed quantum-inspired complex embedding models outperform state-of-the art non-quantum approaches? \\
\item[\textbf{RQ2.}] Out of the two proposed model in this study, which one performs better? \\
\end{itemize}

Table~\ref{table:3} presents the classification accuracy values of all models experimented in this paper, where the bold values indicate the best-performing models for each dataset. It can be clearly seen from the table that CE-Mix is the best-performing model, because it occupies the highest accuracy value on 4 out of 5 benchmarking datasets, and on the remaining dataset it performs only slightly worse than the best-performed model. 

In order to make the results more convincing, we also conduct two-tailed p-tests on the performances. The hypotheses are:

\begin{itemize}
\item[\textbf{H0.}] There is no difference between two groups of performances on a particular dataset.\\
\item[\textbf{H1.}] There is a difference between two groups of performances on a particular dataset.\\
\end{itemize}

We use the threshold 0.05 to accept or reject the null hypothesis: when the obtained p-value $<$ 0.05, the null hypothesis is rejected; when p-value $>$0.05, the null hypothesis is accepted.

Regarding RQ1, it can be observed that CE-Sup and CE-Mix achieves consistently higher or comparable accuracy than non-quantum models under experiment. It illustrates the superiority of complex embedding network over traditional language model (Unigram-TFIDF) (p-value $<$ 0.05 on all datasets, rejecting the null hypothesis, and so forth), unsupervised embeddings trained from external corpus (word2vec, fastText) (p-value $<$ 0.05 on all datasets except MPQA), as well as supervised embedding methods (CaptionRep, DictRep and Paragram-Phrase) (p-value$<$ on all datasets except MPQA). The fair comparison with real embedding network (p-value $<$ 0.05 on all datasets) confirms the superiority of complex embedding over real embedding techniques. 

Regarding RQ2, out of the two complex embedding models proposed in this study, CE-Mix performs consistently but insignificantly (p-value $>$ 0.05) better than CE-Sup in all datasets. Even though it is yet a fully convincing evidence, this result provides us with some intuition that it seems better to model a sentence as a classical mixture of word projectors rather than as a superposition state of latent concepts. For future work we will evaluate the performances of these two models on other datasets as well as other tasks to reach a more solid conclusion.

\section{Conclusion and Future Work}

This paper attempts to address the challenge of representing the combinatory meaning of words for word embedding. The successful applications of quantum-based models in IR tasks inspires us to construct Hilbert Space representation of words and sentences, and explore to build two quantum models for solving sentence classification task. The experimental result on five benchmarking datasets demonstrates their effectiveness.

This work contributes to the fields of both word embeddings and quantum-inspired IR. On the one hand, our work can be interpreted as an improved embedding approach, which tackles the challenge of capturing the emergent meaning of a combination of words. On the other hand, this can be viewed as a pioneering study on quantum-inspired language models with complex numbers, and also an trial effort to adopt the theoretical QWeb framework onto an application context.

For future work, it is necessary to conduct a more comprehensive evaluation of the proposed models, either by evaluating on more datasets or by evaluating the qualities of the trained complex embeddings. We are also looking forward to seek additional ways to model a sentence based on the word states, and the application of the models onto other NLP tasks.

\section*{ACKNOWLEDGEMENT}
This work is supported by the Quantum Access and Retrieval Theory (QUARTZ) project, which has received funding from the European Union's Horizon 2020 research and innovation programme under the Marie Skłodowska-Curie grant agreement No. 721321.

\bibliographystyle{acl_natbib}
\bibliography{acl2018}

\end{document}